\documentclass[12pt]{article}
\usepackage{graphicx}
\usepackage[marginal]{footmisc}
\usepackage{subfigure}
\usepackage{multirow}
\usepackage[noadjust]{cite}
\usepackage{amsmath,amsthm}
\usepackage{amssymb,amsfonts}
\usepackage{caption}
\usepackage{cuted}  
\usepackage{authblk}
\usepackage{booktabs}
\usepackage{url}

\newcommand{\tabincell}[2]{\begin{tabular}{@{}#1@{}}#2\end{tabular}}

\newtheorem{definition}{\textbf{Definition}}

\newtheorem{theorem}{\textbf{Theorem}}

\newtheorem{corollary}{\textbf{Corollary}}

\title{Discovering Association with Copula Entropy}
\author{Ma Jian\thanks{Email: majian@hitachi.cn}}
\affil{\normalsize Hitachi (China) Research \& Development Corporation}

\begin{document}

\maketitle

\begin{abstract}
	\noindent
	Discovering associations is of central importance in scientific practices. Currently, most researches consider only linear association measured by correlation coefficient, which has its theoretical limitations. In this paper, we propose a method for discovering association with copula entropy -- a universally applicable association measure for not only linear cases, but nonlinear cases. The advantage of the method based on copula entropy over traditional method is demonstrated on the NHANES data by discovering more biomedical meaningful associations.
\end{abstract}


\section{Introduction}
\label{s:introduction}
\subsection{Association in sciences}
\noindent
In empirical sciences, researchers collect data from real world systems for making scientific discoveries with statistical tools. Association (or dependence) is such a statistical tool defined for measuring the relationships between random variables of real systems \cite{1}. Correlation, as the linear version of association, is the most commonly considered one in real applications, while statistical dependence covers more broad types of associations including nonlinear cases than correlation does. Another closely related concept, Causality is defined for causal relationships in physical, social and biological systems. Even it is well known that association does not imply causation, association is still a necessary condition for causality in general.

Association and causality are of significant importance in healthcare and medicine \cite{1,2}. In medical research, association is widely used as first evidence for scientific discoveries. Causality is much fundamental in all branches of medicine -- clinicians diagnose based on symptom-disease relationships, pharmacologists find drugs according to drugs' effect on disease, epidemiology study how environmental factors affect population healthcare, etc.. Therefore, association discovery and causal inference are enduring topics in medical research. This paper focuses on association.

\subsection{Traditional association measures}
\noindent
In statistics, classical assication measures have their birth at the early days of the discipline. The most widely used association measure is called "Correlation Coefficient" (CC), proposed by Pearson, and hence also called "Pearson's correlation coefficient" \cite{pearson1897}. It is defined for bi-variate distribution by dividing the covariance of two random variables by the product of their standard deviations, as follows:
\begin{equation}
\rho_{XY} = corr(X,Y) = \frac{cov(X,Y)}{\delta_{X}\delta_{Y}}
\end{equation}
where $X,Y$ represent random variables, $cov$ represents for covariance, and $\delta$ represents for standard deviation. Pearson correlation coefficient is a parametric version of CC. There are also rank-based non-parametric version of CC named after the inventors: Spearman's $\rho$ \cite{spearman1904} and Kendall's $\tau$ \cite{kendall1938}.

\subsection{Association discovery}
\noindent
Association discovery is a common practice in scientific researches. Scientists usually try to discover association -- the statistical relationship between random variables from data with association measures. The associations such discovered can be as evidences for new knowledge and insights. Traditional measure considered in most research has its limitations. 

Copula Entropy (CE) is a rigorously defined mathematical concept proposed for measuring statistical independence \cite{4}. It enjoys many theoretical properties which traditional measures don't have. In this paper, we propose a method for discovering association with CE. CE based association measure has been applied to study the rainfall-runoff relationship in hydrology \cite{5,6}, brain connectivity in neuroscience \cite{7}, weather influence on renewable energy systems \cite{8}. In this paper, it will be used to analyze biomedical data for discovering meaningful associations to demonstrate its advantages over traditional method with CC.

\section{Copula Entropy}
\label{s:CopEnt}
\subsection{Theory}
\noindent
Copula theory unifies representation of multivariate dependence with copula function \cite{9,10}. According to Sklar theorem \cite{11}, multivariate density function can be represented as a product of its marginals and copula density function which represents dependence structure among random variables. This section is to define an association measure with copula. For clarity, please refer to \cite{4} for notations.

With copula density, Copula Entropy is define as follows \cite{4}:
\begin{definition}[Copula Entropy]
\label{d:ce}
	Let $\mathbf{X}$ be random variables with marginals $\mathbf{u}$ and copula density $c(\mathbf{u})$. CE of $\mathbf{X}$ is defined as
	\begin{equation}
	H_c(\mathbf{X})=-\int_{\mathbf{u}}{c(\mathbf{u})\log{c(\mathbf{u})}}d\mathbf{u}.
	\label{eq:ce}
	\end{equation}
\end{definition}

In information theory, Mutual Information (MI) and entropy are two different concepts \cite{12}. In \cite{4}, Ma and Sun proved that MI is actually a kind of entropy, negative CE, stated as follows: 
\begin{theorem}
\label{thm1}
	MI of random variables is equivalent to negative CE:
	\begin{equation}
	I(\mathbf{X})=-H_c(\mathbf{X}).
	\end{equation}
\end{theorem}

Theorem \ref{thm1} has simple proof \cite{4} and an instant corollary (Corollary \ref{c:ce}) on the relationship between information containing in joint probability density function, marginals and copula density.
\begin{corollary}
\label{c:ce}
	\begin{equation}
		H(\mathbf{X})=\sum_{i}{H(X_i)}+H_c(\mathbf{X})
	\end{equation}
\end{corollary}
The above results cast insight into the relationship between entropy, MI, and copula through CE, and therefore build a bridge between information theory and copula theory. CE itself provides a theoretical concept of statistical independence measure.

\subsection{Estimation}
\label{s:est}
\noindent
It is widely considered that estimating MI is notoriously difficult. Under the blessing of Theorem \ref{thm1}, Ma and Sun \cite{4} proposed a non-parametric method for estimating CE (MI) from data which composes of only two steps: \footnote{The code is available at \url{https://github.com/majianthu/copent}.}
\begin{enumerate}
	\item Estimating Empirical Copula Density (ECD);
	\item Estimating CE.
\end{enumerate}

For Step 1, if given data samples $\{\mathbf{x}_1,\ldots,\mathbf{x}_T\}$ i.i.d. generated from random variables $\mathbf{X}=\{x_1,\ldots,x_N\}^T$, one can easily estimate ECD as follows:
\begin{equation}
F_i(x_i)=\frac{1}{T}\sum_{t=1}^{T}{\chi(\mathbf{x}_{t}^{i}\leq x_i)},
\end{equation}
where $i=1,\ldots,N$ and $\chi$ represents for indicator function. Let $\mathbf{u}=[F_1,\ldots,F_N]$, and then one can derives a new samples set $\{\mathbf{u}_1,\ldots,\mathbf{u}_T\}$ as data from ECD $c(\mathbf{u})$. 

Once ECD is estimated, Step 2 is essentially a problem of entropy estimation which can be tackled by many existing methods. Among those methods, the kNN method \cite{13} was suggested in \cite{4}, which leads to a non-parametric way of estimating CE.

\section{CE as association measure}
\noindent
In Section \ref{s:CopEnt}, CE is defined as a measure of statistical dependence with copula function which contains all the dependence information between random variables. Rigorously defined, CE has several properties which an ideal statistical independence measure should have, including multivariate, symmetric, non-negative (0 iff independent), invariant to monotonic transformations, and equivalent to correlation coefficient in Gaussian cases.

Theoretically, CE has many advantages over traditional association measure -- CC. Implied by definition, CC is a bivariate measure with Gaussian assumption while CE has no such limitation. More theoretical comparisons between CC and CE are listed in Table \ref{t:t1a}. 

\begin{table}
\setlength{\abovecaptionskip}{0pt}
\setlength{\belowcaptionskip}{5pt}
\centering
\caption{Theoretical comparisons between CC and CE.}
\begin{tabular}{c|c|c}
	\toprule
	&CC&CE\\
	\midrule
	Linearity&Linear&linear/Non-linear\\
	Order&Second&All\\
	Assumption&Gaussian&None\\
	Dimensions&bivariate&multivariate\\
	Association Type&correlation&dependence\\
	\bottomrule
\end{tabular}
\label{t:t1a}
\end{table}

CC has its two non-parametric version -- Spearman's $\rho$ and Kendall's $\tau$, both of which can be expressed with copula. the former can be represented as:
\begin{equation}
\rho_{XY}=12\int_{u}\int_{v}C(u,v)dudv - 3,
\label{eq:spearman}
\end{equation}
where $X,Y$ are random variables, and $u,v$ are their marginals. The latter can be expressed with copula as follows:
\begin{equation}
\tau_{XY}=4\int_{u}\int_{v}C(u,v)dC(u,v) - 1,
\label{eq:kendall}
\end{equation}
where $C$ is the copula of $(X,Y)$. Compared \eqref{eq:ce} with \eqref{eq:spearman} and \eqref{eq:kendall}, one can learn that CE measures all order of independence within copula function while Spearman's $\rho$ and Kendall's $\tau$ measure only the first order of copula function.

Since CE shows clear advantages over CC, we propose a method for discovering association with it. For the method, we suggest estimating CE with non-parametric method in Section \ref{s:est} so that it will make the proposed method universally applicable without make theoretical assumptions on the underlying systems.

\section{Experiments on the NHANES data}
\subsection{Data}
\noindent
We demonstrate the power of the proposed method on the famous NHANES data. The experimental data were collected from the US National Health and Nutrition Examination Surveys (NHANES) 2013-2014 \cite{14}. The NHANES target population is the non-institutionalized civilian resident population of the United States. The major objectives of NHANES are to monitor trends and emerging issues of population health and to investigate its relationship with risk factors, nutritions and environmental exposures, etc..

During 2013-2014, 14,332 persons from 30 different survey locations were selected for NHANES. Of those selected, 10,175 completed the interview and 9,813 were examined. The collected data are of 5 groups: demographics, dietary, examination, laboratory, and questionnaire. The NHANES collected biological specimens for laboratory analysis to provide detailed information about participants' health and nutritional status. The specimens collected in NHANES 2013-2014 include: blood, urine, oral rinse, and vaginal swabs \cite{15,16}. In this research, the laboratory data is used, which includes 423 variables, some of them with missing values \cite{14}. Discovering the associations between these biological variables can help to understand the data better.

\subsection{Experiments}
\noindent
Three experiments were conducted on laboratory data with two association measures, CC (Pearson's $r$ and Spearman's $\rho$) and CE respectively. Each analysis produced an association matrix, of which each element is the association strength between a pair of variables in the dataset. If a group of variables are highly associated, then the corresponding elements in matrix will have relatively high value simultaneously. Association will be discovered according to the matrix and then explained by domain knowledge. Comparison between CC and CE will be made.

In the experiments, CE was estimated with two step non-parametric method in Section \ref{s:est}. The missing values were filled with the mean of their corresponding variables.

\subsection{Results}
\begin{figure*}
	\centering
\subfigure[Pearson's $r$]{\includegraphics[width=0.45\textwidth]{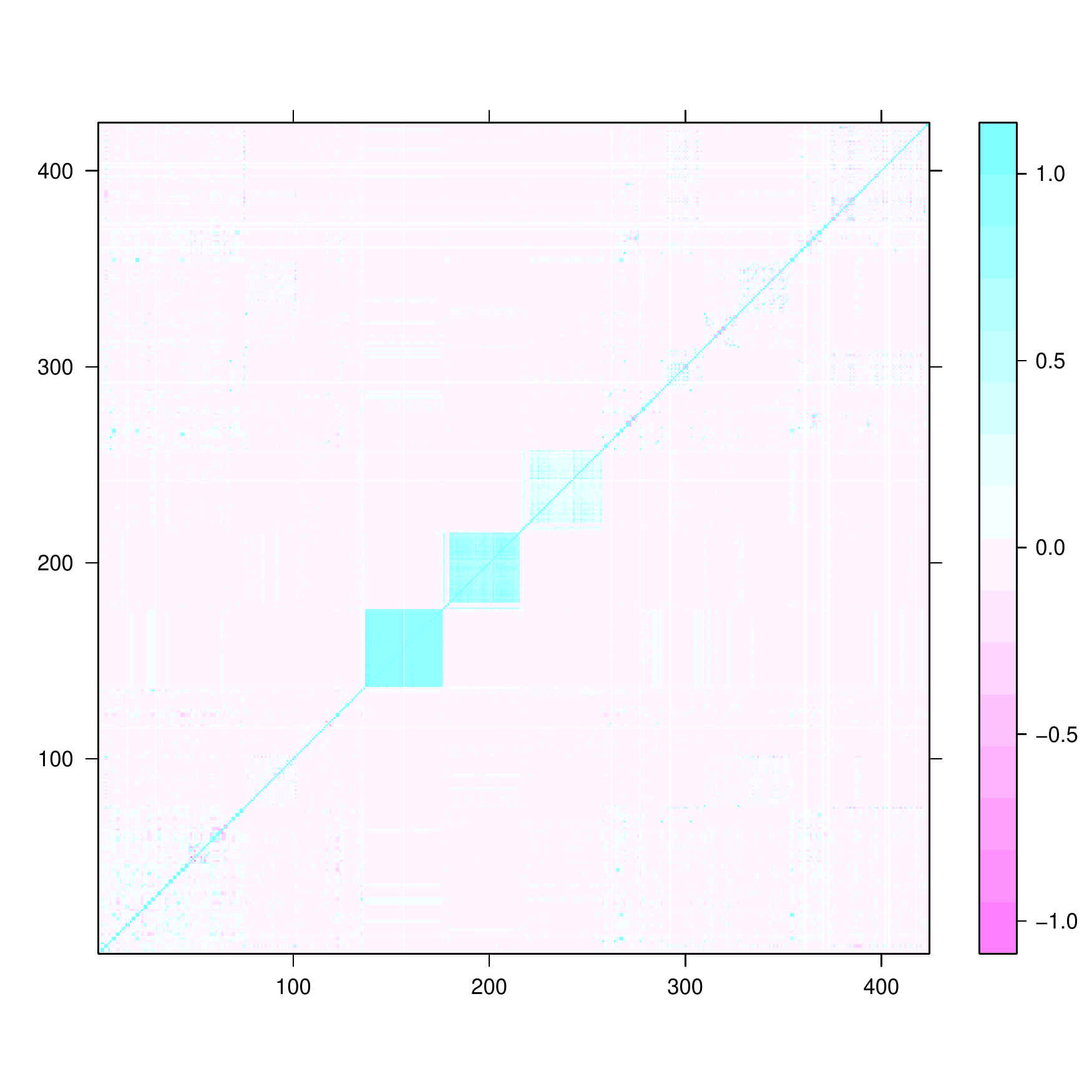}	
	\label{f:cor1}}
\hfil
\subfigure[Spearman's $\rho$]{\includegraphics[width=0.45\textwidth]{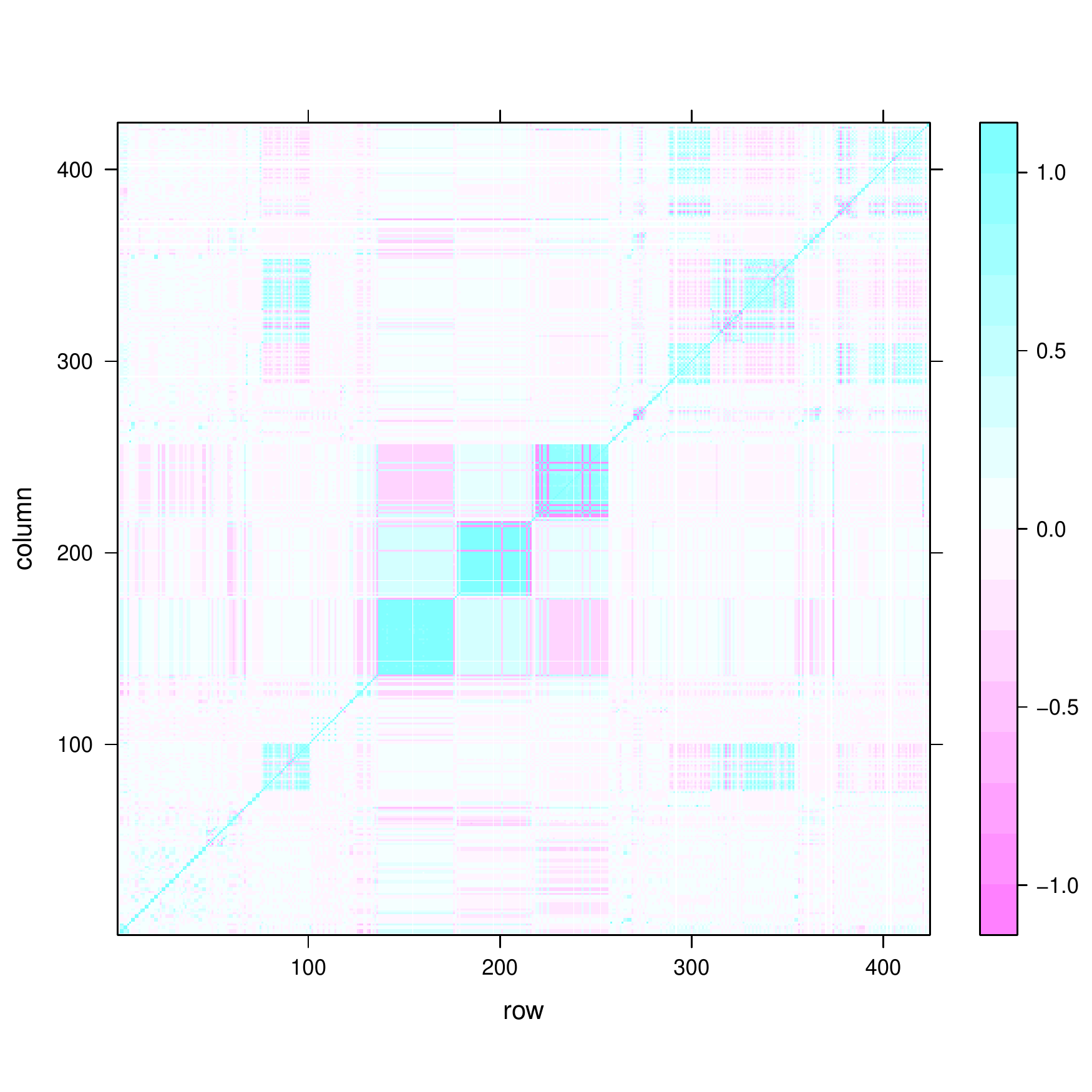}	
	\label{f:cor2}}
\hfil
\subfigure[CE]{\includegraphics[width=0.65\textwidth]{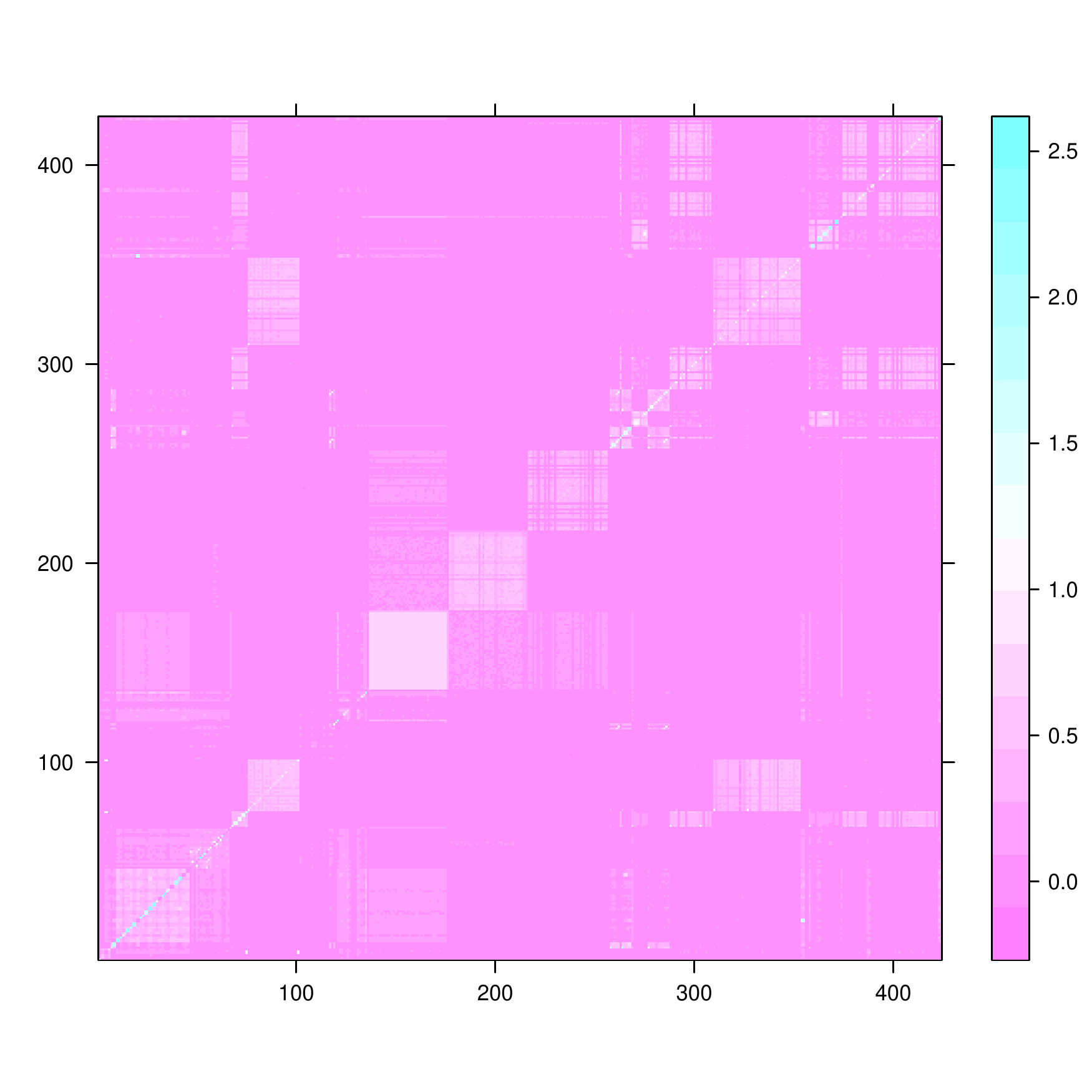}	
	\label{f:mi}}
\caption{Association matrices of the laboratory data: (a) Pearson's $r$, (b) Spearman's $\rho$, (c) CE.}
\label{f:figure1}
\end{figure*}

\noindent
Experimental results on laboratory data are shown in Figure \ref{f:figure1}. It can be easily learned from the association matrix that CE based association matrix shows a much clearer picture of how the laboratory testing variables correlated into different groups than CC based association matrix do. Pearson's $r$ presents 3 groups of correlated variables on the diagonal of its matrix which are also presented in the CE matrix and Spearman's $\rho$ also presents a better result with more correlation identified. Meanwhile, The CE matrix presents more associations on off the diagonal parts of the matrix than the Pearson CC matrix does. According to the matrix, the associations by CE are identified into 5 groups as listed in Table \ref{t:t2}.

\section{Discussion}
\noindent
It can be learned from Figure \ref{f:figure1} that CE presents much clearer correlation pattern than two CC measures do. It is because that CE measures all order of statistical dependence while CC measures only second order of that. Figure \ref{f:figure1} shows that Spearman's $\rho$ presents the result much closer to the result by CE than Pearson's $r$ does because it is also estimated non-parametrically. The theoretical difference between CE and Spearman's $\rho$ is that they have different representation of copula function. In this sense, one can learn that CE captures a more clear picture of statistical dependence than Spearman's $\rho$ does when comparing Figure \ref{f:mi} with Figure \ref{f:cor2}, which provide a evidence that CE has theoretical advantages over two CC measures. A clear difference is that the former presents positive association strength of all order dependence while the latter presents relative strength between [-1,1] of second order dependence. In a word, the results suggest that CE captures more information of assication relationship than CCs do.

Association relations between laboratory testing variables discovered by CE based analysis in Table \ref{t:t2} can be well explained by biomedical knowledge. The explanation for association of the variables of group 2 is straightforward -- they measure the different biochemical elements in bloods, which should be correlated obviously. 

The variables of group 3 (Oral Glucose, Insulin and Cholesterol) are used to test the risk of type II diabetes. The level of Triglycerides indicates how well body turns food into energy. Insulin protects glucose -- energy source inside human body. Insulin also allows human body to use triglycerides for energy. As a indicator of Type II diabetes risk, high Triglycerides signal insulin resistance -- excess insulin and Glucose. In this sense, group 3 shows the biological picture of diabetes \cite{17,18}. 

The variables of group 1 are related to human exposure to environmental chemicals. Polycyclic Aromatic Hydrocarbons (PAH) are a group of about 100 different chemicals that are usually generated from traffic-related air pollutant, cooking pollutant and smoking. PAH and metals co-exists in nature and are considered together in study of effect on health of environmental exposure \cite{19,20}. 

The variables of group 5 are related to daily life consumption of plastics. Phthalates are a group of chemicals which are widely used in the production of plastics to improve the material quality. They are also called Plasticizers. All the associated variables in group 5 are closely related to the wide spread human exposure to plastic made consumer products \cite{21,22}.

\begin{table*}
\setlength{\abovecaptionskip}{0pt}
\setlength{\belowcaptionskip}{5pt}
\centering \footnotesize
\caption{Associated variables groups found by CE based analysis.}
\begin{tabular}{c|c|c} 
	\toprule
	Group&Index&Laboratory Variables \\
	\midrule
	\multirow{3}*{1} &288-302&Polycyclic Aromatic Hydrocarbons (PAH) - Urine\\
					\cline{2-3}
					&68-75&Copper, Selenium \& Zinc - Serum \\
					\cline{2-3}
					&395-420&Urine Metals\\
	\hline
	\multirow{2}*{2} &358-373&Blood Lead, Cadmium, Total Mercury, Selenium, and Manganese\\
					\cline{2-3}
					&269-276&Blood mercury: inorganic, ethyl and methyl \\
	\hline	
	\multirow{3}*{3} &277-287&Oral Glucose Tolerance Test\\
					\cline{2-3}
					&258-262&Insulin \\
					\cline{2-3}
					&7-9&\tabincell{c}{Cholesterol - LDL, Triglyceride \& Apoliprotein (ApoB),\\
					WTSAF2YR- Fasting Subsample 2 Year MEC Weight, \\
					LBXAPB - Apolipoprotein (B) (mg/dL), \\
					LBDAPBSI - Apolipoprotein (B) (g/L)}\\
	\hline
	\multirow{2}*{4} &10-46&Standard Biochemistry Profile\\
					\cline{2-3}
					&137-176&Human Papillomavirus (HPV) - Oral Rinse \\
	\hline	
	\multirow{2}*{5} &76-101&Personal Care and Consumer Product Chemicals and Metabolites\\
					\cline{2-3}
					&327-353&Phthalates and Plasticizers Metabolites - Urine \\
	\bottomrule	
\end{tabular}
\label{t:t2}
\end{table*}

\section{Conclusion}
\noindent
We propose a method for discovering association with copula entropy. We also compare CE with CC on several theoretical aspects and conclude that the former is more theoretically advanced and much widely applicable than the latter. The advantage of the proposed method based on CE over traditional one is demonstrated on the NHANES data by discovering more biomedical meaningful associations. It is believed that the method based on CE has great potential on making more scientific discoveries in its physical, social and biological applications.

\vskip 2mm
\noindent
\textbf{Acknowledgement}
\vskip 2mm

\small
\noindent
The author thanks Matsumori Masaki for comments and suggestions.

\vskip 2mm

\renewcommand\refname{

\end{document}